\documentclass[sigconf, screen]{acmart}

\usepackage{balance}
\usepackage{iftex}
\ifPDFTeX
\usepackage[utf8]{inputenc}
\usepackage[T1]{fontenc}
\else
\ifXeTeX
\usepackage{fontspec}
\else 
\usepackage{luatextra}
\fi
\fi

\usepackage{cjhebrew}
\AtBeginDocument{%
	\providecommand\BibTeX{{%
			\normalfont B\kern-0.5em{\scshape i\kern-0.25em b}\kern-0.8em\TeX}}}

\usepackage{subcaption}
\usepackage{graphicx}
\usepackage{colortbl}
\usepackage{hyperref}

\setcopyright{acmcopyright}

\begin{document}

	\copyrightyear{2019}
	\acmYear{2019}
	\acmConference[SUMAC '19]{1st Workshop on Structuring and Understanding of Multimedia heritAge Contents}{October 21, 2019}{Nice, France}
	\acmBooktitle{1st Workshop on Structuring and Understanding of Multimedia heritAge Contents (SUMAC '19), October 21, 2019, Nice, France}
	\acmPrice{15.00}
	\acmDOI{10.1145/3347317.3357242}
	\acmISBN{978-1-4503-6910-7/19/10}

	\fancyhead{}
	% do not delete this code.

	%%
	%% The "title" command has an optional parameter,
	%% allowing the author to define a "short title" to be used in page headers.
	%\title{Gender Recognition in Art History using Deep Learning}
	%\title{Character Recognition in Art History Using Deep Learning}
	% VC: Problem: OCR ambiguity -> Figure Recognition? Or how about this?
	%\title{Art Character Recognition in Historical Artworks Using Deep Learning}
	\title{Recognizing Characters in Art History Using Deep Learning}
	%%
	%% The "author" command and its associated commands are used to define
	%% the authors and their affiliations.
	%% Of note is the shared affiliation of the first two authors, and the
	%% "authornote" and "authornotemark" commands
	%% used to denote shared contribution to the research.
	\author{Prathmesh Madhu}
	\affiliation{\institution{Pattern Recognition Lab, Friedrich-Alexander-Universität Erlangen-Nürnberg, Germany}}
	\email{prathmesh.madhu@fau.de}
	\author{Ronak Kosti}
	\affiliation{\institution{Pattern Recognition Lab, Friedrich-Alexander-Universität Erlangen-Nürnberg, Germany}}
	\email{ronak.kosti@fau.de}
	\author{Lara Mührenberg}
	\affiliation{\institution{Institute of Church History, Friedrich-Alexander-Universität Erlangen-Nürnberg, Germany}}
	\email{lara.muehrenberg@fau.de}
	\author{Peter Bell}
	\affiliation{\institution{Institute for Art History, Friedrich-Alexander-Universität Erlangen-Nürnberg, Germany}}
	\email{peter.bell@fau.de}
	\author{Andreas Maier}
	\affiliation{\institution{Pattern Recognition Lab, Friedrich-Alexander-Universität Erlangen-Nürnberg, Germany}}
	\email{andreas.maier@fau.de}
	\author{Vincent Christlein}
	\affiliation{\institution{Pattern Recognition Lab, Friedrich-Alexander-Universität Erlangen-Nürnberg, Germany}}
	\email{vincent.christlein@fau.de}
	% 	\author{Prathmesh Madhu}
	% 	\authornotemark[4]
	% 	\authornotemark[1]
	% 	\author{Ronak Kosti}
	% 	\authornotemark[4]
	% 	\authornote{Both authors contributed equally to this research}
	% 	%\orcid{1234-5678-9012}
	% 	\author{Lara Mührenberg}
	% 	\authornote{Institute of Church History}
	% 	\author{Peter Bell}
	% 	\authornote{Institute for Art History}
	% 	\author{Andreas Maier}
	% 	\authornotemark[4]
	% 	\author{Vincent Christlein}
	% 	\authornote{Pattern Recognition Lab}
	% 	\email{{prathmesh.madhu, ronak.kosti, lara.muehrenberg, peter.bell, andreas.maier, vincent.christlein}@fau.de}
	% 	\affiliation{%
	% 		\institution{Friedrich-Alexander-Universität Erlangen-Nürnberg, Germany}
	% 		%		\streetaddress{Martenstrasse 3}
	% 		%		\city{Erlangen}
	% 		%		\state{Bavaria}
	% 		%		\postcode{91052}
	% 	}
	\renewcommand{\shortauthors}{Madhu and Kosti, et al.}
	
	%\author{Lars Th{\o}rv{\"a}ld}
	%\affiliation{%
	%  \institution{The Th{\o}rv{\"a}ld Group}
	%  \streetaddress{1 Th{\o}rv{\"a}ld Circle}
	%  \city{Hekla}
	%  \country{Iceland}}
	%\email{larst@affiliation.org}
	%
	%\author{Valerie B\'eranger}
	%\affiliation{%
	%  \institution{Inria Paris-Rocquencourt}
	%  \city{Rocquencourt}
	%  \country{France}
	%}
	%
	%\author{Aparna Patel}
	%\affiliation{%
	% \institution{Rajiv Gandhi University}
	% \streetaddress{Rono-Hills}
	% \city{Doimukh}
	% \state{Arunachal Pradesh}
	% \country{India}}
	%
	%\author{Huifen Chan}
	%\affiliation{%
	%  \institution{Tsinghua University}
	%  \streetaddress{30 Shuangqing Rd}
	%  \city{Haidian Qu}
	%  \state{Beijing Shi}
	%  \country{China}}
	%
	%\author{Charles Palmer}
	%\affiliation{%
	%  \institution{Palmer Research Laboratories}
	%  \streetaddress{8600 Datapoint Drive}
	%  \city{San Antonio}
	%  \state{Texas}
	%  \postcode{78229}}
	%\email{cpalmer@prl.com}
	%
	%\author{John Smith}
	%\affiliation{\institution{The Th{\o}rv{\"a}ld Group}}
	%\email{jsmith@affiliation.org}
	%
	%\author{Julius P. Kumquat}
	%\affiliation{\institution{The Kumquat Consortium}}
	%\email{jpkumquat@consortium.net}
	
	%%
	%% By default, the full list of authors will be used in the page
	%% headers. Often, this list is too long, and will overlap
	%% other information printed in the page headers. This command allows
	%% the author to define a more concise list
	%% of authors' names for this purpose.
	\renewcommand{\shortauthors}{Madhu and Kosti, et al.}
	
	%%
	%% The abstract is a short summary of the work to be presented in the
	%% article.
	\begin{abstract}
		%\rk{+ gender detection in digital humanities, specifically art history (with mary and gabriel as the main proponents since they sometimes show ambiguous genders).}
		%\\
		%\rk{+ importance of using CNNs - a powerful visual features extractor. Learns from Big data}

		In the field of Art History, images of artworks and their contexts are core to understanding the underlying semantic information. However, the highly complex and sophisticated representation of these artworks makes it difficult, even for the experts, to analyze the scene. From the computer vision perspective, the task of analyzing such artworks can be divided into sub-problems by taking a bottom-up approach. In this paper, we focus on the problem of recognizing the characters in Art History. From the iconography of \emph{Annunciation of the Lord} (Figure~\ref{fig:teaser}), we consider the representation of the main protagonists, \emph{Mary} and \emph{Gabriel}, across different artworks and styles. We investigate and present the findings of training a character classifier on features extracted from their face images. The limitations of this method, and the inherent ambiguity in the representation of \emph{Gabriel}, motivated us to consider their bodies (a bigger context) to analyze in order to recognize the characters. Convolutional Neural Networks (CNN) trained on the bodies of \emph{Mary} and \emph{Gabriel} are able to learn person related features and ultimately improve the performance of character recognition. We introduce a new technique that generates more data with similar styles, effectively creating data in the similar domain. We present experiments and analysis on three different models and show that the model trained on domain related data gives the best performance for recognizing character. Additionally, we analyze the localized image regions for the network predictions. Code is open-sourced and available at \href{https://github.com/prathmeshrmadhu/recognize_characters_art_history}{this https URL} and the link to the published peer-reviewed article is \href{https://dl.acm.org/citation.cfm?id=3357242}{this dl acm library link}.
		%In the field of Art History, semantic understanding of paintings and artworks is of utmost importance to the experts. In order to reach the semantics of the scene - the iconography - one can divide the problem into mini sub-problems like gender and age detection, indoor or outdoor scene, pose estimates of the person in the scene and so on. In this paper, we target the gender detection problem in the art history sub-field in cooperation with the digital humanities, specifically on the "Annunciation of the Lord" scene. We investigate and present the drawbacks of training the classifier on only facial images which motivates to train the classifier on the body images of Mary and Gabriel (two main protagonists of annunciation scene). Deep networks based on convolutional neural networks have shown tremendous success in recent times in the area of feature representation, learning and solving the problems like classification and object detection. Our classifier network is a variant of Resnet50~\cite{resnet50paper}.\\
		\keywords{Recognizing Characters  \and Machine Learning \and Convolutional Neural Networks.}
	\end{abstract}
%	\keywords{Recognizing Characters  \and Machine Learning \and Convolutional Neural Networks.}
	
	\begin{CCSXML}
		<ccs2012>
		<concept>
		<concept_id>10010147.10010178.10010224</concept_id>
		<concept_desc>Computing methodologies~Computer vision</concept_desc>
		<concept_significance>500</concept_significance>
		</concept>
		<concept>
		<concept_id>10010147.10010257</concept_id>
		<concept_desc>Computing methodologies~Machine learning</concept_desc>
		<concept_significance>500</concept_significance>
		</concept>
		<concept>
		<concept_id>10010405.10010469</concept_id>
		<concept_desc>Applied computing~Arts and humanities</concept_desc>
		<concept_significance>500</concept_significance>
		</concept>
		<concept>
		<concept_id>10010147.10010178.10010224.10010240.10010241</concept_id>
		<concept_desc>Computing methodologies~Image representations</concept_desc>
		<concept_significance>500</concept_significance>
		</concept>
		</ccs2012>
	\end{CCSXML}
	
	\ccsdesc[500]{Computing methodologies~Computer vision}
	\ccsdesc[500]{Computing methodologies~Image representations}
	\ccsdesc[500]{Computing methodologies~Machine learning}
	\ccsdesc[500]{Applied computing~Arts and humanities}

	%% A "teaser" image appears between the author and affiliation
	%% information and the body of the document, and typically spans the
	%% page.
	\begin{teaserfigure}
		\centering
		\includegraphics[width=0.49\textwidth,  keepaspectratio]{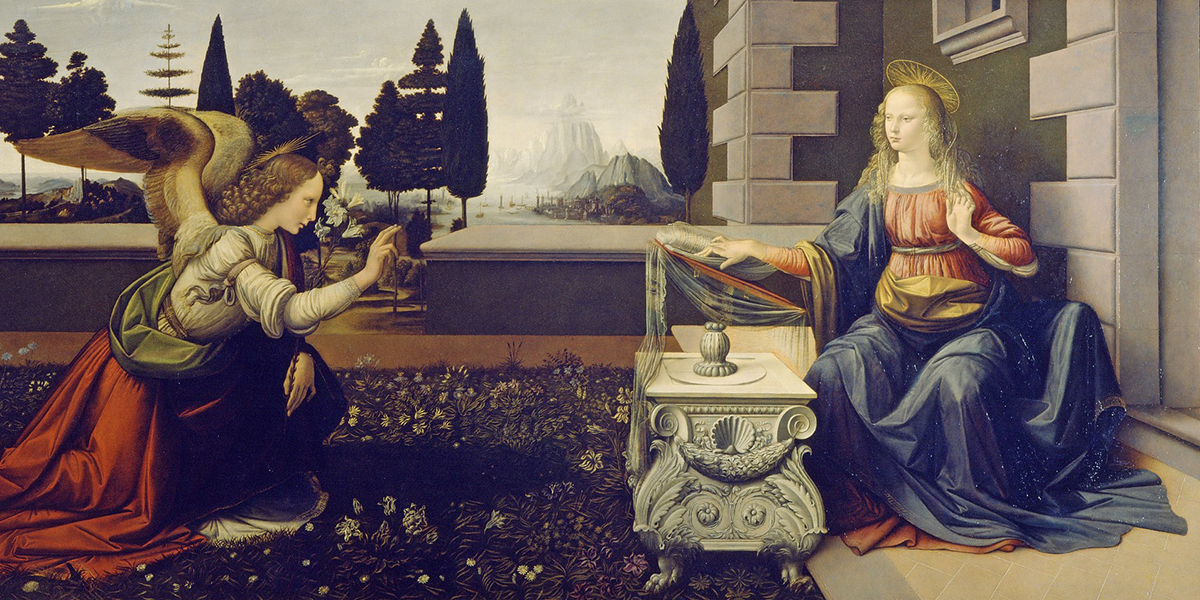}
		\includegraphics[width=0.49\textwidth, height=0.245\textwidth]{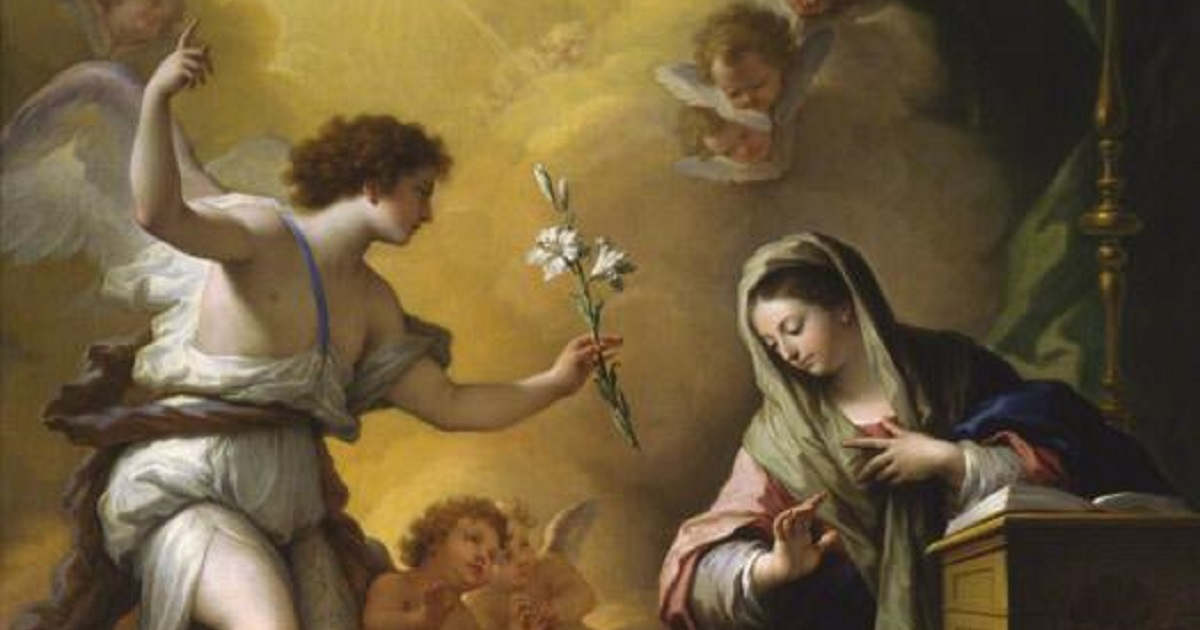}
		\caption{Art historical scene depicting the iconography called \emph{Annunciation of the Lord} (left \cite{vinci}, right \cite{paolo}). Mary and Gabriel are the main protagonists. We can clearly see the differences in the background, in the artistic style, in the foreground, in the objects, their properties, and the use of color.}
		%\Description{Enjoying the baseball game from the third-base
		%	seats. Ichiro Suzuki preparing to bat.}
		\label{fig:teaser}
	\end{teaserfigure}
	
	%%
	%% This command processes the author and affiliation and title
	%% information and builds the first part of the formatted document.
	\maketitle
	
	%%%%%%%%%%%%%%%%%%%%%%%%%%%%%%%%%%%%%%%%%%%%%%%%%%%%%%%%%%%%%%%%%%
	\section{Introduction}
	% 	\rk{+ introduce the problem of gender detection with a few examples. Specifically explaining about the importance in the field of digital heritage, or in general for interesting interpretations of the story/iconography based on the apparent gender}
	% 	\rk{+ How automatic gender detection will help the current understanding of the gender perspective}
	
\begin{figure}[!t]
		\centering
		%% mary
		\begin{subfigure}[t]{0.24\linewidth}
			\centering
			\includegraphics[width=1.8cm, height=4cm, keepaspectratio]{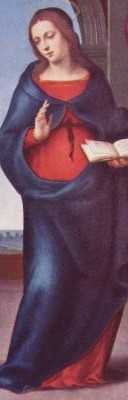}
			\caption{}
		\end{subfigure}% 
		\begin{subfigure}[t]{0.24\linewidth}
			\centering
			\includegraphics[width=1.8cm, height=4cm, keepaspectratio]{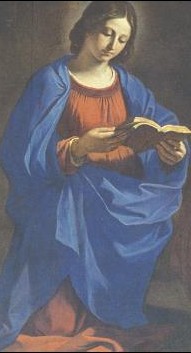}
			\caption{}
		\end{subfigure}%
		\begin{subfigure}[t]{0.24\linewidth}
			\centering
			\includegraphics[width=1.8cm, height=4cm, keepaspectratio]{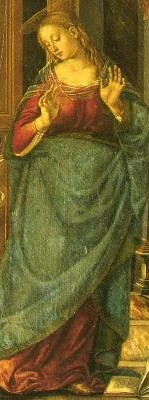}
			\caption{}
		\end{subfigure}%
		\begin{subfigure}[t]{0.24\linewidth}
			\centering
			\includegraphics[width=1.8cm, height=4cm, keepaspectratio ]{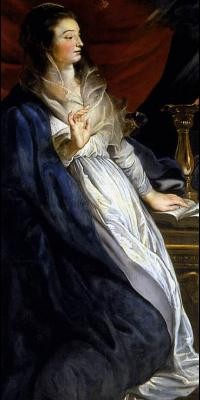}
			\caption{}
		\end{subfigure}
		%% gabriel 
		\begin{subfigure}[t]{0.24\linewidth}
			\centering
			\includegraphics[width=1.8cm, height=4cm, keepaspectratio]{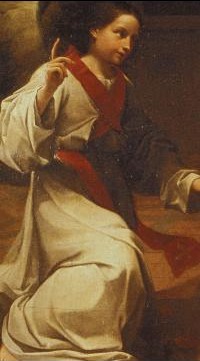}
			\caption{}
		\end{subfigure}% 
		\begin{subfigure}[t]{0.24\linewidth}
			\centering
			\includegraphics[width=1.8cm, height=4cm, keepaspectratio]{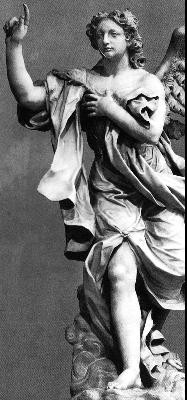}
			\caption{}
		\end{subfigure}%
		\begin{subfigure}[t]{0.24\linewidth}
			\centering
			\includegraphics[width=1.8cm, height=4cm, keepaspectratio]{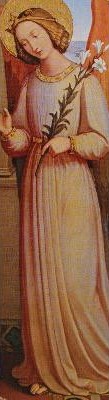}
			\caption{}
		\end{subfigure}%
		\begin{subfigure}[t]{0.24\linewidth}
			\centering
			\includegraphics[width=1.8cm, height=4cm, keepaspectratio]{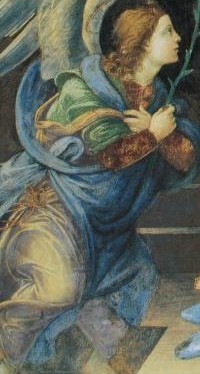}
			\caption{}
		\end{subfigure}
		\caption{Depiction of Mary (a -- d) and Gabriel (e -- f) as the main protagonists in the \emph{Annunciation of the Lord} iconography}
		\label{fig:samples_mary_gabriel}
	\end{figure}

	The understanding of scenes in art images, specifically called as iconography, is a very demanding task due to the complexity of the scene. Interpreting a scene involves detection and recognition of various present objects, their relationship to each other, their impact on the visual perception of the scene and their relevance for the respective task. In images and paintings from various artworks, the understanding of a scene becomes more complex even for an expert for the following reasons: \emph{(1)}, the artworks are an artistic representation of real-life objects, people and scenes or inspired by these; and \emph{(2)} the artistic style differs from one artist to another and it also may change from one painting to another, even for an artwork of a given theme~\cite{crowley2014search, crowley2014state, crowley2015face}. For example, in Figure~\ref{fig:samples_mary_gabriel} we can see Mary (a - d) and Gabriel (e - h) represented by different artists and their respective styles. Interesting to note here is that all of the images are from the same iconography called \emph{Annunciation of the Lord}, however their representation differs from one image to another. Sometimes the artworks are fragmented or have unique compositions. Artists quite often employed different means to convey a message. For example, they would use the body pose of the main protagonists (observe the differences in the body pose of Mary and Gabriel in Figure~\ref{fig:teaser}) or even gestures \cite{bell2013nonverbal, impett2017totentanz}.This makes the interpretation of an artwork more difficult, even for an expert. From computer vision perspective, this problem poses an interesting challenge for vision techniques to understand an artwork with some objectivity. Understanding art involves complex processes such as discovering and recognizing the background structure of the scene, the objects present and their significance within the scene, their relationship with the main object of focus, artistic style and the higher level semantics to deconstruct the meaning of the artwork. With recent advances in computer vision, human-level performance has been achieved in many tasks like detection, recognition and segmentation of objects on natural real world images~\cite{coco}. However, understanding the underlying semantic knowledge still remains an open challenge~\cite{zhou2019semantic}. 
	
	Our work on recognizing character in artworks attempts to explore recognition features about the main subjects depicted in the given scene. This analysis is of great importance for art history. For example, Figure~\ref{fig:samples_mary_gabriel} shows examples of \emph{Mary} and \emph{Gabriel}. They are the main protagonists and hence the center of focus in the iconography called \emph{Annunciation of the Lord}. Figure~\ref{fig:samples_mary_gabriel} (a -- d) shows Mary in the female form, however, the representation of Gabriel (e -- h) is not always clear. In humanities, opinions about Gabriel's gender are divided, with some claiming that it is male while others arguing that it has no gender at all. This is because Angels are regarded as beings created by God, mediating between the heavenly and earthly spheres. The corporeal nature of angels is discussed in detail in the Judeo-Christian tradition~\cite{tavard_1982}. Early Christianity attributed an etheric body to them~\cite{rosenberg_1967}, and in the High Middle Ages, the doctrine prevailed that angels had a purely spiritual body ~\cite{krauss_2001}. Unlike humans, angels do not have a sex or gender. In the biblical stories, the angels appear in human bodies (more precisely male), whose appearance they can assume. They are therefore also called ``sons of God'' or ``sons of heaven". The name of the Archangel Gabriel contains the word stem that means ``man" or ``strength" ~\cite{michel_1962}. In the book of Daniel, his appearance is described as ``like a man" (Dan 8:15). In art, angels are first shown in human male form, analogous to the biblical findings ~\cite{holl_1968}. Their fiery nature (ether) can be made visible by the red colouring of their skin  ~\cite{rosenberg_1967}, their belonging to the heavenly sphere by bluish translucent flesh ~\cite{holl_1968}. From the Middle Ages onward, the asexuality of the bodies of angels was sometimes more strongly depicted ~\cite{klauser_1962}. The angel Gabriel himself therefore has no sex or gender, but manifests himself in a human appearance, which can have male and female parts. These can also be made visible in art. Due to this ambiguity it is very difficult to classify Gabriel as purely male or female. In this work, we aim to solve a sub-problem of recognizing \emph{Mary} and \emph{Gabriel} within artworks. In order to recognize the characters of Mary and Gabriel, we introduce approaches that can accomplish this task within art images, regardless of the type of art, its style or the artist. %Given a query image, recognizing the character within the image can be divided into two core tasks: \emph{(1)} extracting the features related to that person, and \emph{(2)} using these features, classifying the person as one character or the other.
	
	In the last few years, several methods have been introduced for object recognition (including people) on real-world datasets in computer vision. For example, traditional methods using hand-crafted features, such as filter-based feature extractors~\cite{lowe1999object}, which can be aggregated using Bag of (visual) Words~\cite{FeiFei05}. Current state-of-art systems use CNN-based deep networks as object feature extractors~\cite{liu2018path, zhou2014object, dai2017deformable}. However, the protagonists represented in the artworks have very different and unique characteristics compared to photographs (as shown in Figure~\ref{fig:samples_wiki}). Hence, it becomes very important to use the domain knowledge present in the artworks for recognizing character.
	
	The majority of the existing techniques in computer vision~\cite{Rothe-IJCV-2016, eidinger2014age, parkhi2015deep, ranjan2017hyperface} rely on facial features in order to recognize the person. Since the artworks are representations of the artist's imagination and skill, it is difficult to recognize the person from their facial features alone. A case in point would be that of Gabriel. As seen in Figure~\ref{fig:samples_mary_gabriel}(e - h), Gabriel's face gives an impression of being a boy (e), man (f), a woman (g) or the perception is unclear (h). The use of only faces for recognizing character in artworks is therefore insufficient. In our proposed method, we use the entire body images of Mary and Gabriel to model the problem of recognizing character. 
	
	In addition, it also would be interesting to find out the semantic understanding of the trained model when it recognizes a particular character from an artwork. Usually, in artworks, the face is used in addition to other meta-information such as clothing or the neighboring contextual information to analyze the paintings. Figure~\ref{fig:teaser} shows two images from the iconography of \emph{Annunciation of the Lord}. We can see that there is an angel-like figure to the left with a pose pointing to another figure on the right of the image. The figure on the right is that of Mary. It is represented as a female form, while the left figure is that of the Archangel Gabriel, whose representation seems neutral without apparent gender reference. In order to achieve a higher semantic understanding of the scene depicted here, it is important for the vision model to recognize these characters.
	
	In this paper we demonstrate that recognizing specific characters (Mary and Gabriel) is possible using deep models. Our contributions are as follows: \emph{(1)} we show that the performance of traditional machine learning models, such as SVMs (Linear and RBF kernels), Logistic Regression and Random Forests decreases when they are trained on the whole body of the characters as opposed to only faces; \emph{(2)} we introduce a novel technique as a way of simplifying the transfer of knowledge from one domain to another; \emph{(3)} we show that this technique is beneficial for the model's performance and outperforms traditional machine learning algorithms. 
	
	The paper is organized as follows: In section~\ref{sec:relwork}, we discuss about the related work in person identification and transfer learning that could be useful for recognizing characters in art; In section~\ref{sec:met}, we introduce traditional as well as deep learning based methods adopted by us for recognizing characters and also furnish the details about the dataset preparation; Section~\ref{sec:eval} essentially details the experimental setup for all the methods and their corresponding quantitative and qualitative evaluations; and In section~\ref{sec:conclude} we make a small discussion and conclusions about our adopted methods and their respective merits. 
	\section{Related Work}\label{sec:relwork}
	%The gender classification problem can be split into two parts a) feature extractor that extracts features from the face b) feature classifier which assigns a particular class (male / female) to the feature. Various feature extraction methods include use of raw pixel face images [Moghaddam and Yang, 2002], Principal Component Analysis (PCA) [Balci and Atalay, 2002], Independent Component Analysis (ICA) [Jain and Huang, 2004], Local Binary Pattern (LBP) [Yang and Ai, 2007; Shan, 2010; Sun. et al, 2006; Chen and Ross, 2011], Linear Discriminant Analysis (LDA) [Bekios Calfa et al., 2011] and metrology [Cao et al., 2011]. On the other hand, the feature classifier methods include Neural networks [Golomb et al, 1990; Khan et al, 2005], Adaboost [Baluja and Rowly, 2007; Yang and Ai, 2007; Sun et al, 2006; Shan 2010], Gaussian process [Kim et al, 2006] and Support Vector Machines (SVM) [Moghaddam and Yang, 2002; Guo et al, 2009; Chen and Ross, 2011]
	
	Person identification has been a core problem in computer vision. Since the advent of video surveillance technologies, cheap hardware for recording and storing videos and the latest research in object recognition techniques, it has become important to automatically identify a person in an image or a video to keep a track on the movements for security reasons. Many common methods use facial features for the recognition of a person. Usually, the character (or person) identification is divided into two parts: \emph{(1)} feature extraction and \emph{(2)} feature classification that recognizes the character's identity. In traditional computer vision techniques, features are hand-crafted, e.,g., LBP, HOG, SIFT or an ensemble of these local descriptors to encode the information present in the face~\cite{sivic2009you, simonyan2014learning, cinbis2011unsupervised, sivic2005person}. On the other hand, the feature classifier methods include neural networks~\cite{golomb1990sexnet}, Adaboost~\cite{baluja2007boosting} and Support Vector Machines (SVM)~\cite{moghaddam2002learning}. With the advent of the recent end-to-end deep learning algorithms, raw face pixel intensity values are given as input to the CNN-based classifiers for training and then these trained classifiers are used to assign a class to an unseen face image. CNNs use convolutional filters to learn the facial features by training millions of example images~\cite{parkhi2015deep, fan2014learning}. These methods provide a unique advantage of training a network in an end-to-end fashion. 
	
	\begin{table}
		\caption{Face Datasets}
		\label{tab:facedata}
		\begin{tabular}{ccccc}
			\toprule
			& IMDB-Wiki  & LFW-Face  & Adience  & CelebA \\
			\midrule
			Source & Wikipedia & Web & Flickr & Google\\
			\# Images & 62,328 & 13,233 & 26,580 & 202,599\\
			\# Identities & 20,284 & 5749 & 2,284 & 10,177\\
			Reference &~\cite{Rothe-IJCV-2016} &~\cite{huang2008labeled} &~\cite{eidinger2014age} &~\cite{liu2015faceattributes} \\
			%			\midrule
			%			\O & 1 in 1,000& For Swedish names\\
			%			$\pi$ & 1 in 5& Common in math\\
			%			\$ & 4 in 5 & Used in business\\
			%			$\Psi^2_1$ & 1 in 40,000& Unexplained usage\\
			\bottomrule
		\end{tabular}
	\end{table}
	
	The implementation of deep models was particularly successful due to the availability of large datasets. Table~\ref{tab:facedata} shows the details of available face datasets with their respective sources, number of images, and number of identities. The number of images typically needed to train a CNN from scratch is huge and in the range of millions~\cite{krizhevsky2012imagenet}. However, the face datasets (Table~\ref{tab:facedata}) do not have such large amounts of data. This situation poses a dilemma: Since the deep networks need large amounts of data to train from scratch, the available face data is insufficient in comparison.
	
	Transfer learning is a method where the knowledge learned in one domain can be carefully transferred to another, which can be beneficial~\cite{Yin_2019_CVPR, pan2009survey}. Typically, the new domain to be adapted for transfer learning must have a similar feature space and data distribution as the original domain. The current deep networks that are trained on person or face recognition have used transfer learning by using pre-trained networks to jump start the training~\cite{Zhong_2019_CVPR, parkhi2015deep, Rothe-ICCVW-2015}. These pre-trained networks provide a good starting point for training without requiring large amounts of data.

	\begin{figure}[!t]
		\centering
		%% original
		\begin{subfigure}[t]{0.24\linewidth}
			\centering
			\includegraphics[width=1.8cm, height=4cm, keepaspectratio]{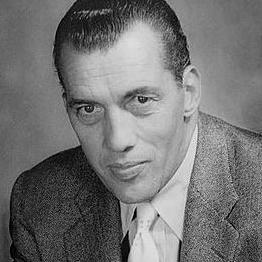}
		\end{subfigure} 
		\begin{subfigure}[t]{0.24\linewidth}
			\centering
			\includegraphics[width=1.8cm, height=4cm, keepaspectratio]{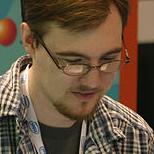}
		\end{subfigure}%
		\begin{subfigure}[t]{0.24\linewidth}
			\centering
			\includegraphics[width=1.8cm, height=4cm, keepaspectratio]{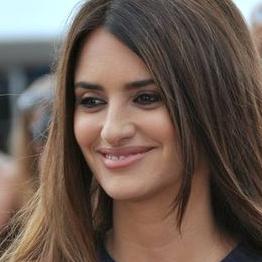}
		\end{subfigure}%
		\begin{subfigure}[t]{0.24\linewidth}
			\centering
			\includegraphics[width=1.8cm, height=4cm, keepaspectratio ]{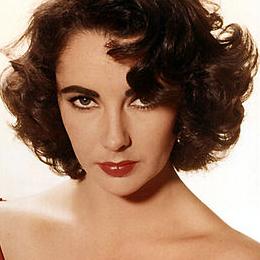}
		\end{subfigure}
		
		%% styled 
		\begin{subfigure}[t]{0.24\linewidth}
			\centering
			\includegraphics[width=1.8cm, height=4cm, keepaspectratio]{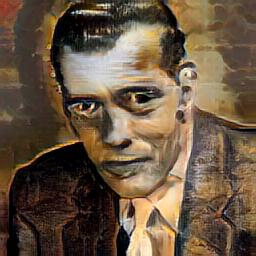}
		\end{subfigure}% 
		\begin{subfigure}[t]{0.24\linewidth}
			\centering
			\includegraphics[width=1.8cm, height=4cm, keepaspectratio]{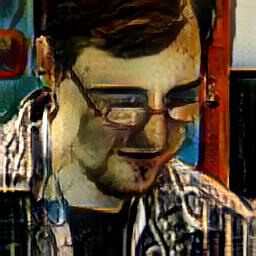}
		\end{subfigure}%
		\begin{subfigure}[t]{0.24\linewidth}
			\centering
			\includegraphics[width=1.8cm, height=4cm, keepaspectratio]{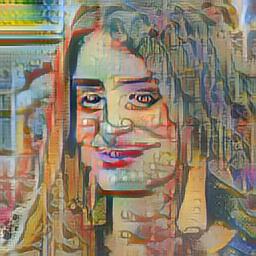}
		\end{subfigure}%
		\begin{subfigure}[t]{0.24\linewidth}
			\centering
			\includegraphics[width=1.8cm, height=4cm, keepaspectratio]{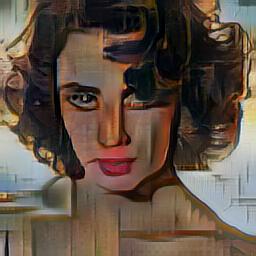}
			
		\end{subfigure}
		
		%% style 
		\begin{subfigure}[t]{0.24\linewidth}
			\centering
			\includegraphics[width=1.8cm, height=4cm, keepaspectratio]{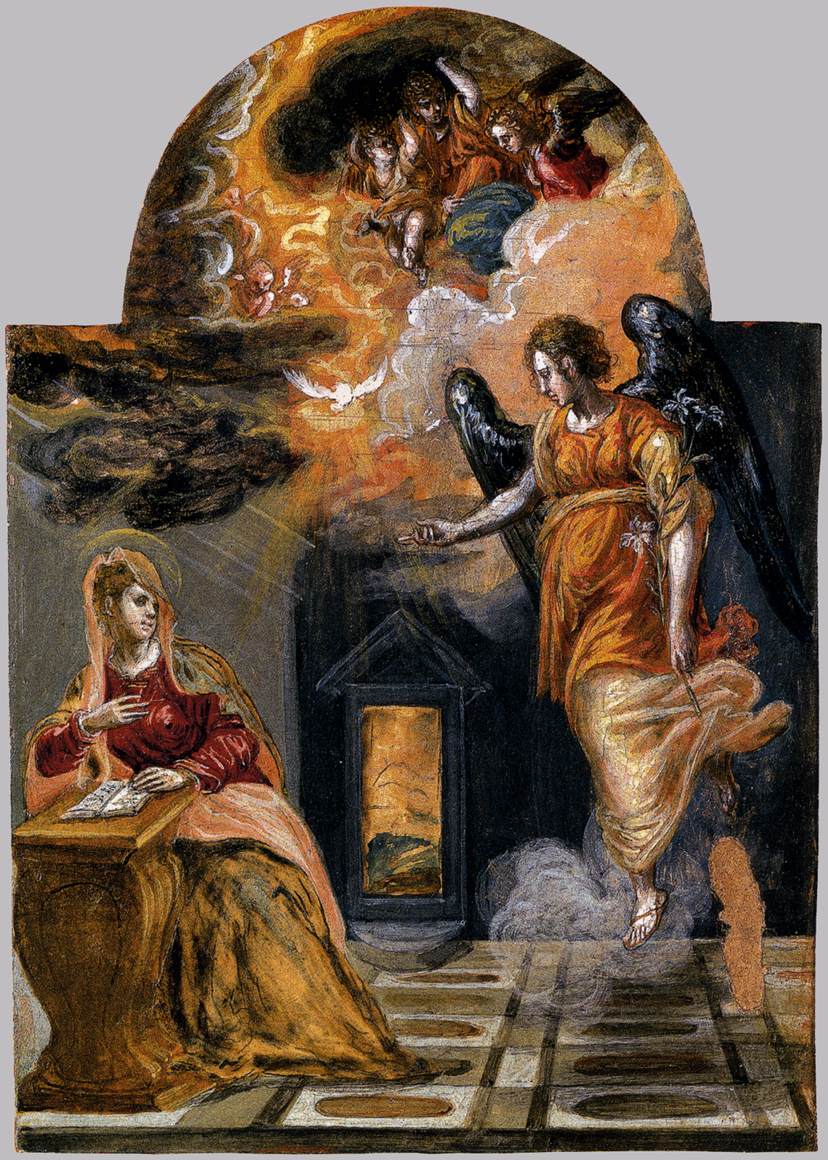}
			\caption{}
		\end{subfigure}% 
		\begin{subfigure}[t]{0.24\linewidth}
			\centering
			\includegraphics[width=1.8cm, height=4cm, keepaspectratio]{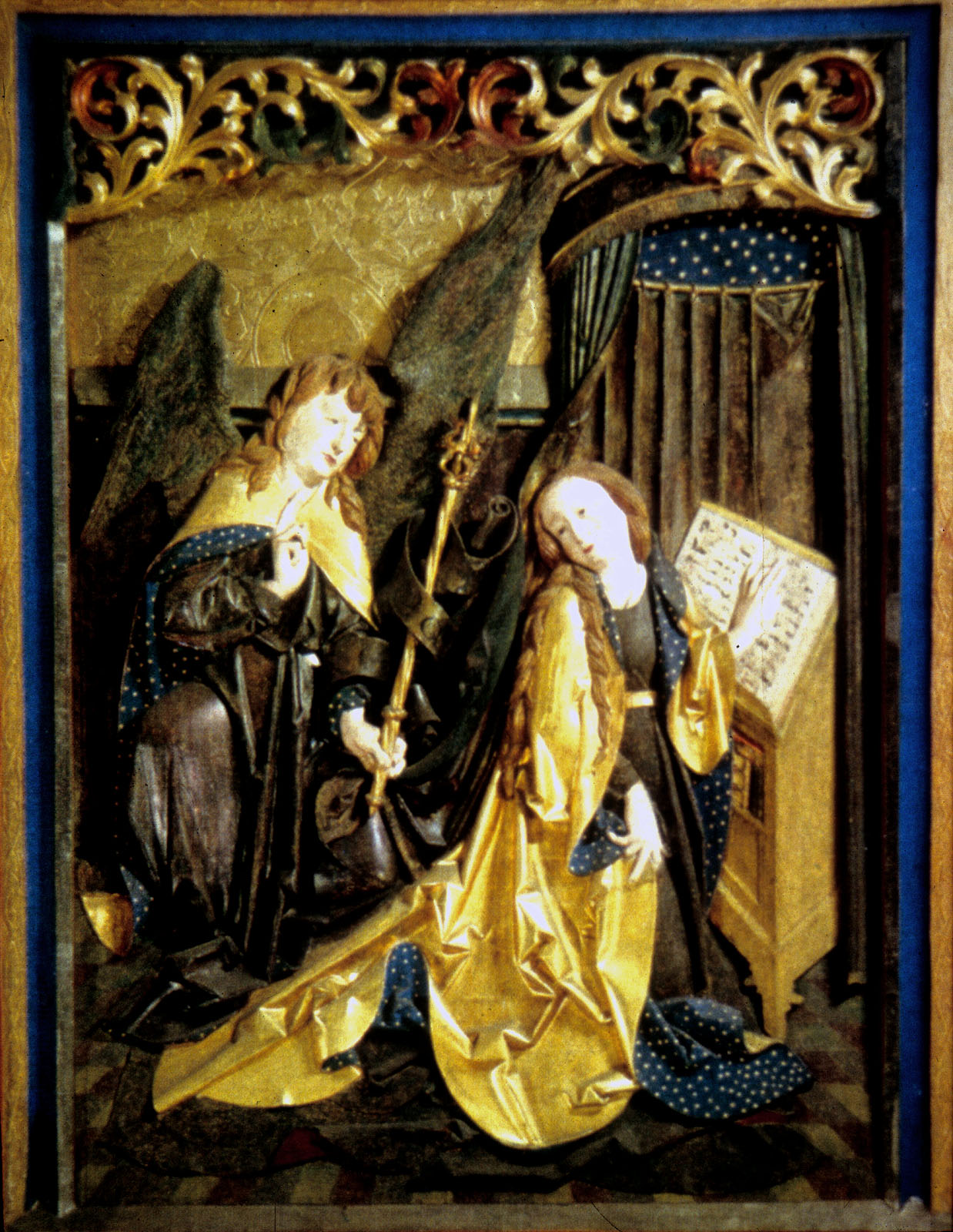}			\caption{}
		\end{subfigure}%
		\begin{subfigure}[t]{0.24\linewidth}
			\centering
			\includegraphics[width=1.8cm, height=4cm, keepaspectratio]{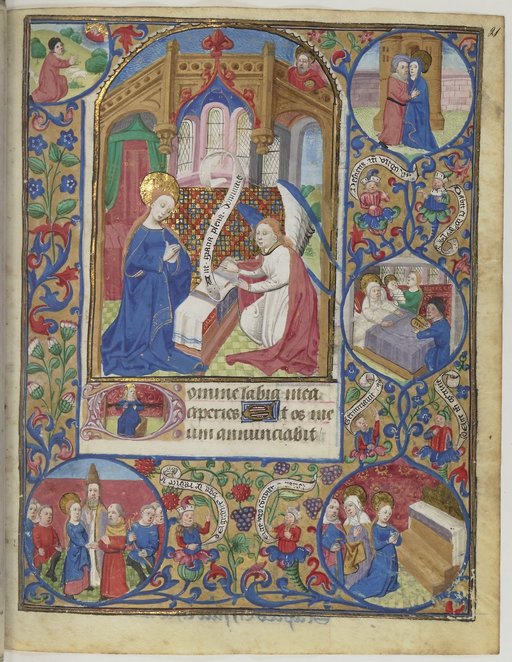}
			\caption{}
		\end{subfigure}%
		\begin{subfigure}[t]{0.24\linewidth}
			\centering
			\includegraphics[width=1.8cm, height=4cm, keepaspectratio]{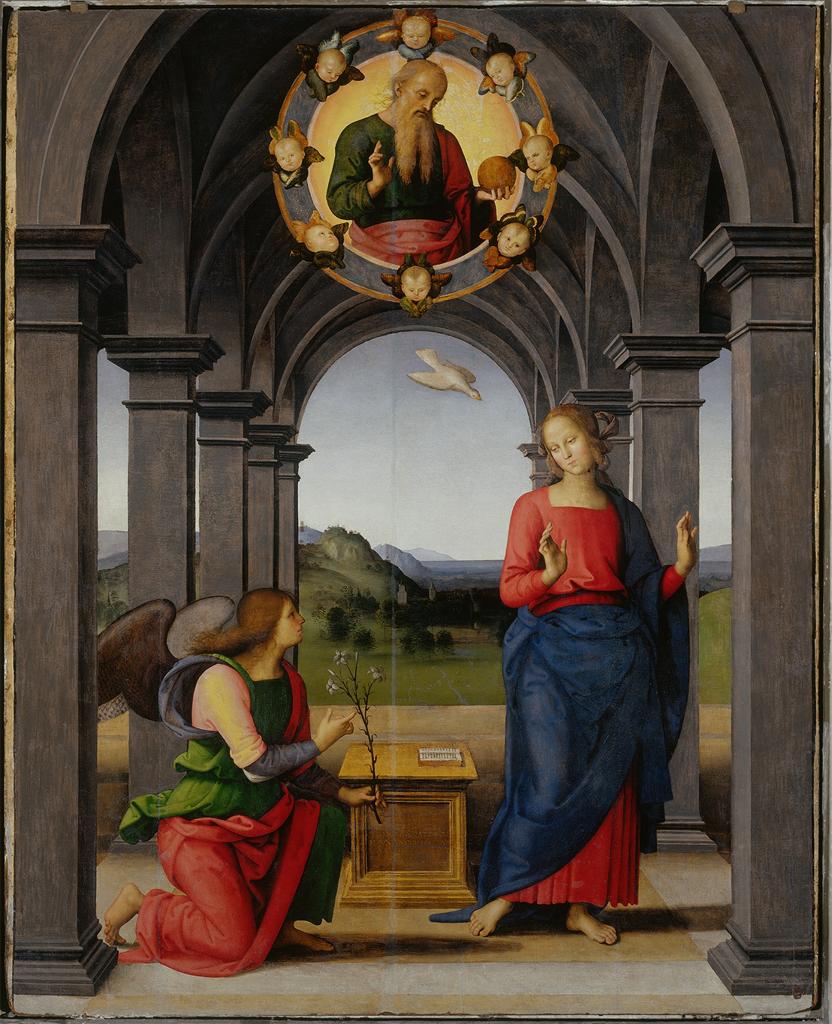}
			\caption{}	
		\end{subfigure}	
		\caption{Face images taken from~\cite{Rothe-IJCV-2016,eidinger2014age} in row~1 with their style-transferred counterparts in row~2 and their corresponding style images in row~3. Columns (a) \& (b) depict images of men whereas columns (c) \& (d) depict women.}
		\label{fig:samples_wiki}
	\end{figure}

	Recent works applying modern computer vision methods to the problems related to artworks claim that the images or paintings they work with differ substantially in semantics and representations from natural images~\cite{crowley2014search, crowley2014state, crowley2015face}. For example, Figure~\ref{fig:samples_wiki} shows sample images from IMDB-Wiki \& Adience datasets~\cite{Rothe-IJCV-2016,eidinger2014age} (row~1). In comparison to Figure~\ref{fig:samples_mary_gabriel}, it is apparent that the artworks show a very different character representation. They could be inspired from real-life or often a manifestation of the artist's imagination. 
	
	Recent research on the semantic understanding of paintings in art, however,  shows that it is possible to apply 
	%vision techniques of multi-modal learning and text retrieval to transfer domain knowledge 
	transfer learning of vision to artworks while training new models. For example, Garcia and Vogiatzis~\cite{garcia2018read} considered the Text2Art challenge where they successfully retrieved the related artwork from the set of test images $45.5\%$ (Top-$10$) of times. They also showed a technique to store visual and textual information of the same artwork in the same semantic space, thereby making the task of retrieval much easier. In another instance, Zhong et al.~\cite{Zhong_2019_CVPR} showed that it is possible to model a deep feature network on a particular artwork collection using self-supervised learning for discovering near-duplicate patterns in larger collections. 
	
	Strezoski and Worring~\cite{Strezoski2017} also showed that using multi-task learning techniques with deep networks for feature learning gives better performance over hand-crafted features. Clearly, it is possible to use techniques like transfer learning from state-of-art in computer vision to adapt for art related problems.

	%%%%%%%%%%%%%%%%%%%%%%%%%%%%%%%%%%%%%%%%%%%%%%%%%%%%%%%%%%%%%%%%%%
	\section{Methodology}\label{sec:met} 
	In this section, we explain how we created our database of annunciation scenes and its related style-transferred dataset. Afterwards, the models used to train on this dataset are described.
	\subsection{Database Creation}
	Our dataset\footnote{data acquired for non-commercial scientific research} consists $2787$ images~\cite{lehre}~\cite{data1}~\cite{data1}~\cite{data2}~\cite{data3}~\cite{data4} of artworks from the iconography called \emph{Annunciation of the Lord}, with focus on the main protagonists: \emph{Mary} and \emph{Gabriel}. The image data is from a corpus of medieval and early modern annunciations, acquired from public domain. We generated bounding box estimates for the bodies of both using a fast object detector called YOLO~\cite{yolov3}. These were corrected manually by art history experts. The distribution of Mary and Gabriel is nearly balanced with $1172$ and $1007$ images, respectively. We also generated bounding boxes for their faces using an image annotation tool called VIA~\cite{dutta2019vgg}. This database served as a source for all the experiments mentioned in section~\ref{subsec:props}, except \textsc{VGGFace-B}.
	
	Table~\ref{tab:facedata} shows some currently available datasets in computer vision for identity recognition of real-life individuals. We chose to use IMDB-Wiki~\cite{eidinger2014age} and Adience~\cite{Rothe-IJCV-2016} since the images are not only limited to faces, but also contain the upper body part. Combining both, we have around $20000$ images belonging to each class, male and female, which we call the \emph{content images}. Since there are fewer female samples, we have chosen a similar number of male samples. Figure~\ref{fig:samples_wiki} (row~1) shows sample images of two men and two women from the combined dataset. Our annunciation dataset has $2787$ images which we call the \emph{style images}. Figure~\ref{fig:samples_wiki} (row~3) shows four sample annunciation scenes serving as style images, from our database. Using a style transfer model, based on adaptive instance normalization, introduced by Huang and Belongie~\cite{huang2017arbitrary}, we transferred the artistic style of \textit{style images} to the \textit{content images}. For each style image, we transfer its corresponding style to $16$ ($8$ males and $8$ females) randomly selected content images. In this way, we are able to use each individual style of the annunciation scenes, resulting in a similar distribution of styles in the style-transferred images. These images have a similar style to the annunciation scene, effectively making the distribution of our content data similar to that of our style data (annunciation scenes). Figure~\ref{fig:samples_wiki} (row~2) shows some samples of the style-transferred images. The styles of row~3 have been transferred to the corresponding images in row~1. Since the content images have categorical labels of \emph{male} and \emph{female}, we now have a similar styled, larger dataset, which can be used to learn the domain knowledge of the annunciation scenes. This database served as a source for training the \textsc{VGGFace-B} model (see below). 
	
	For the annunciation dataset, we used $200$ test images, $100$ each for Mary and Gabriel and the rest for training, making a split of approximately $90\%/10\%$ between train/test. For the styled dataset, we used a split of $75\%/25\%$ between train/test.
	
	In the next subsection, we introduce methods used for training on these datasets. 
	
	\subsection{Proposed Methods}\label{subsec:props}
	In all our proposed methods, we use a ResNet50 model~\cite{he2016deep}, pre-trained on the VGGFace dataset~\cite{parkhi2015deep}. This model was trained to identify people from their face images, so it's feature space is very well trained for recognizing characters.
	
	\begin{figure}[!t]
		\begin{subfigure}{\linewidth}
			\centering
			\includegraphics[width=0.6\linewidth]{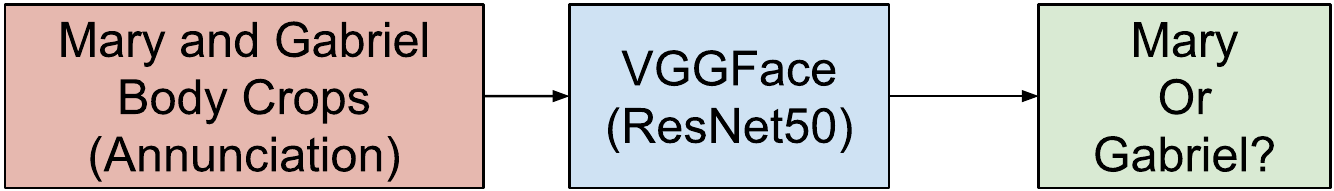}\hfill
			\caption{VGGFace-A}
		\end{subfigure}\par\medskip
		\begin{subfigure}{\linewidth}
			\includegraphics[width=\linewidth]{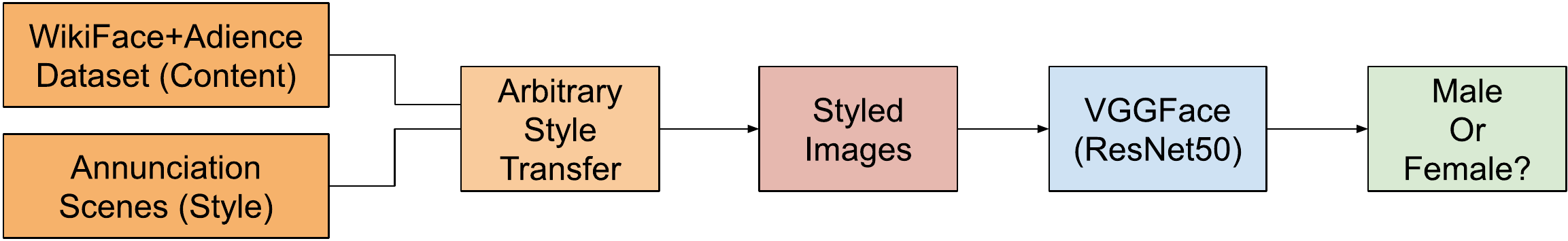}\hfill
			\caption{VGGFace-B}
		\end{subfigure}\par\medskip
		\begin{subfigure}{\linewidth}
			\centering
			\includegraphics[width=0.6\linewidth]{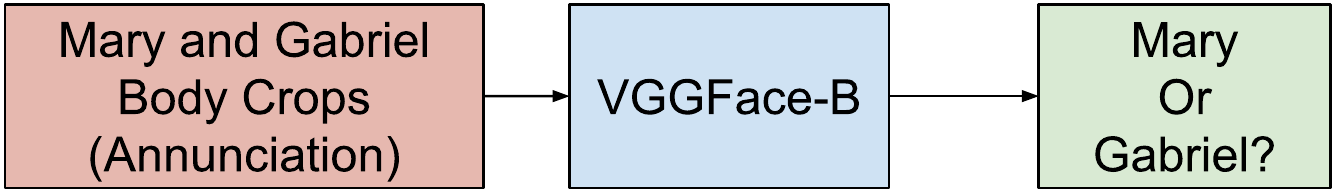}\hfill
			\caption{VGGFace-C}
		\end{subfigure}\par\medskip
		\caption{Pipelines for Recognizing Characters. The model definitions of VGGFace-* are in section~\ref{subsec:props}}
		\label{fig:cha_reg}
	\end{figure}
	
	For recognizing characters, we experiment with the following five methods:	\begin{enumerate}
		\item \textsc{ML-Face:} We take a pretrained ResNet50~\cite{he2016deep}, trained on object recognition and remove the top-most softmax layer. Thus, we use the activations of the penultimate layer ($2048$ dimensional vector) as our features. These are extracted for all the face images of Mary and Gabriel. Using these as input features, we train Random Forests, SVMs (Linear and RBF kernels) and Logistic Regression.
		
		\item \textsc{ML-Body:} The examples of Figure~\ref{fig:samples_mary_gabriel} show the differences in the perception of Mary and Gabriel. The face does not have all the information to perceive the differences between Mary and Gabriel. Hence, we propose to include more contextual information in the form of their bodies. For this experiment, we take the bodies of Mary and Gabriel and extract their ResNet50 features similar to ML-Face.
		
		\item \textsc{VGGFace-A:} For this experiment, we also take a pretrained ResNet50~\cite{he2016deep} trained on the VGGFace dataset~\cite{parkhi2015deep} and replace the top-most $1000$-class softmax layer with a $2$-class sigmoid layer. Since Gabriel's gender is ambiguous, we used sigmoid for final layer activations. The receptive fields of CNNs are able to learn the contextual and hierarchical information~\cite{zhou2016learning}, we fine-tune this model on the body images of Mary and Gabriel. Figure~\ref{fig:cha_reg}\,(a) shows a visual representation of the model.
		
		\item \textsc{VGGFace-B:} Once more, we take a pretrained Resnet50, similar to the previous methods. However, we directly train it on the styled image dataset. Instead of Mary and Gabriel, we take Female and Male as the corresponding labels since we expect the model to learn the styles rather than person related features. Figure~\ref{fig:cha_reg}\,(b) shows a visual representation of the model.
		
		\item \textsc{VGGFace-C:} We take \textsc{VGGFace-B} from the above method and fine-tune it on the Mary and Gabriel bodies dataset. We expect that \textsc{VGGFace-C} will be able to learn specific features related to Mary and Gabriel because VGGFace-B has been trained on an annunciation style related dataset. Figure~\ref{fig:cha_reg}\,(c) shows a visual representation of the model.
		
	\end{enumerate}
	
	%%%%%%%%%%%%%%%%%%%%%%%%%%%%%%%%%%%%%%%%%%%%%%%%%%%%%%%%%%%%%%%%%%
	\section{Evaluation}\label{sec:eval}
	In this section we explain the experiments conducted, their training steps, results and analysis for all the models introduced in subsection~\ref{subsec:props}. All the results in the form of tables or otherwise have been calculated on the images of the test set, i.\,e., the models have not been trained on any image of the test set.  The CNN models were trained by using standard hyperparameter values for learning rates, dropout rates, batch sizes and epochs. We augment the data on the fly with techniques like shear, shift, horizontal flip and rotation. These techniques ensure that the model is not biased towards specific poses or orientation of the characters. 
	
	\begin{table}[!t]
		%\centering
		\caption{Performance metrics for traditional machine learning models trained on the features from \emph{FACE} images of Mary and Gabriel.}
		\begin{tabular}{cllll}
			\toprule
			\textbf{Model Type} & \textbf{Pr} & \textbf{Re} & \textbf{F1} & \textbf{Acc.} \\ 
			\midrule
			Random Forests (200 est.) & 0.75 & 0.75 & 0.75 & 0.75 \\ 
			Logistic Regression & 0.70 & 0.68 & 0.68 & 0.68 \\
			SVM (Linear, C = 100) & 0.78 & 0.78 & 0.78 & 0.78 \\ 
			SVM (RBF, C = 1000, $\gamma$ = 0.01) & \textbf{0.80} & \textbf{0.79} & \textbf{0.79} & \textbf{0.79} \\ 
			\bottomrule
			\label{tab:results_ml}
		\end{tabular}
	\end{table}
	
	\subsection{Experiments with ML-Face and ML-Body}
	We trained these models with features extracted from their respective faces and bodies of Mary and Gabriel. For each of these models, we did a grid search for finding the best parameter for tuning the performance of the models. Table~\ref{tab:results_ml} and Table~\ref{tab:results_all} show the results of these methods for face and body features, respectively. We can see that their performance is not consistent between both the features. In general, the performance is better with face features in comparison to body features. Furthermore, we see that for face features as well as for body features, SVMs show the highest accuracies. The drop in performance of these methods, when face features are compared to body features, indicate that they are unable to process data with higher visual complexity (or contextual content).
	
	\begin{table}[!t]
		%\centering
		\caption{Performance metrics for different models trained on the features from \emph{BODY} images of Mary and Gabriel. The model definitions of VGGFace-* are given in section~\ref{sec:eval}.}
		\begin{tabular}{cllll}
			\toprule
			\textbf{Model Type} & \textbf{Pr} & \textbf{Re} & \textbf{F1} & \textbf{Acc.} \\ 
			\midrule
			Random Forests (200 est.) & 0.59 & 0.56 & 0.54 & 0.59 \\ 
			Logistic Regression & 0.68 & 0.68 & 0.68 & 0.69 \\ 
			SVM (Linear, C = 10) & 0.68 & 0.68 & 0.68 & 0.68 \\ 
			SVM (RBF, C = 1000, $\gamma$ = 0.01) & 0.70 & 0.70 & 0.71 & 0.71 \\ 
			VGGFace-A & 0.77 & 0.70 & 0.73 & 0.72 \\ 
			VGGFace-B & 0.53 & 0.49 & 0.51 & 0.49 \\ 
			\textbf{VGGFace-C} & \textbf{0.84} & \textbf{0.76} & \textbf{0.79} & \textbf{0.79} \\ \bottomrule
			\label{tab:results_all}
		\end{tabular}
	\end{table}
	
	\subsection{Experiments with VGGFace-* Models} 
	All of these models are pretrained VGGFace~\cite{parkhi2015deep} models (pretrained ResNet50 on the VGGFace dataset), hence we use a low learning rate of $1e-4$ with batch size of $32$ for fine-tuning. Training is done on a randomly generated split of training and validation set~\cite{sklearn_api} of images from our dataset of Mary and Gabriel. 
	
\begin{figure}[!t]
		\begin{subfigure}{0.45\linewidth}
			\centering
			\includegraphics[width=\linewidth]{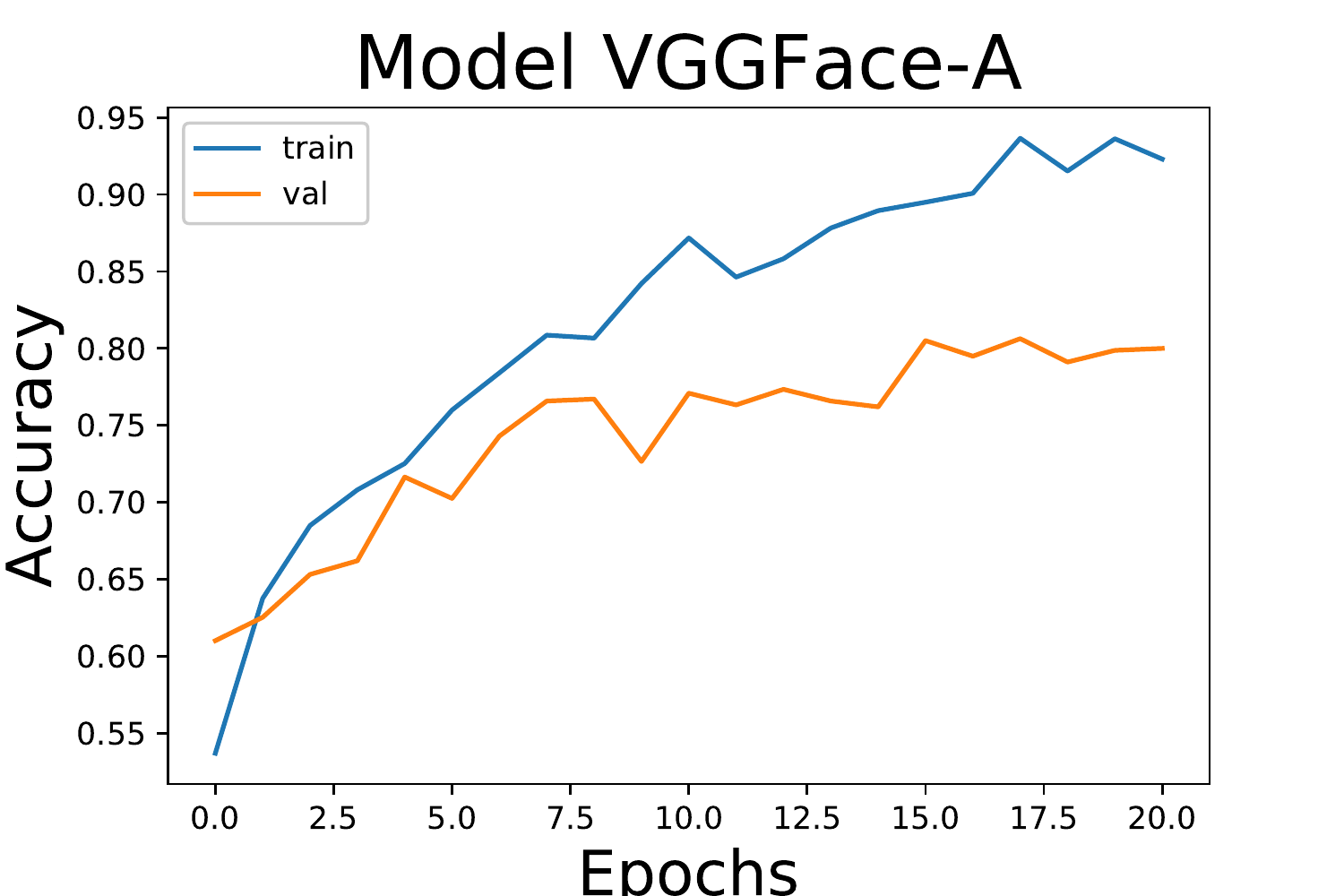}
		\end{subfigure}	\vspace{4pt}	
		\begin{subfigure}{0.45\linewidth}
			\centering
			\includegraphics[width=\linewidth]{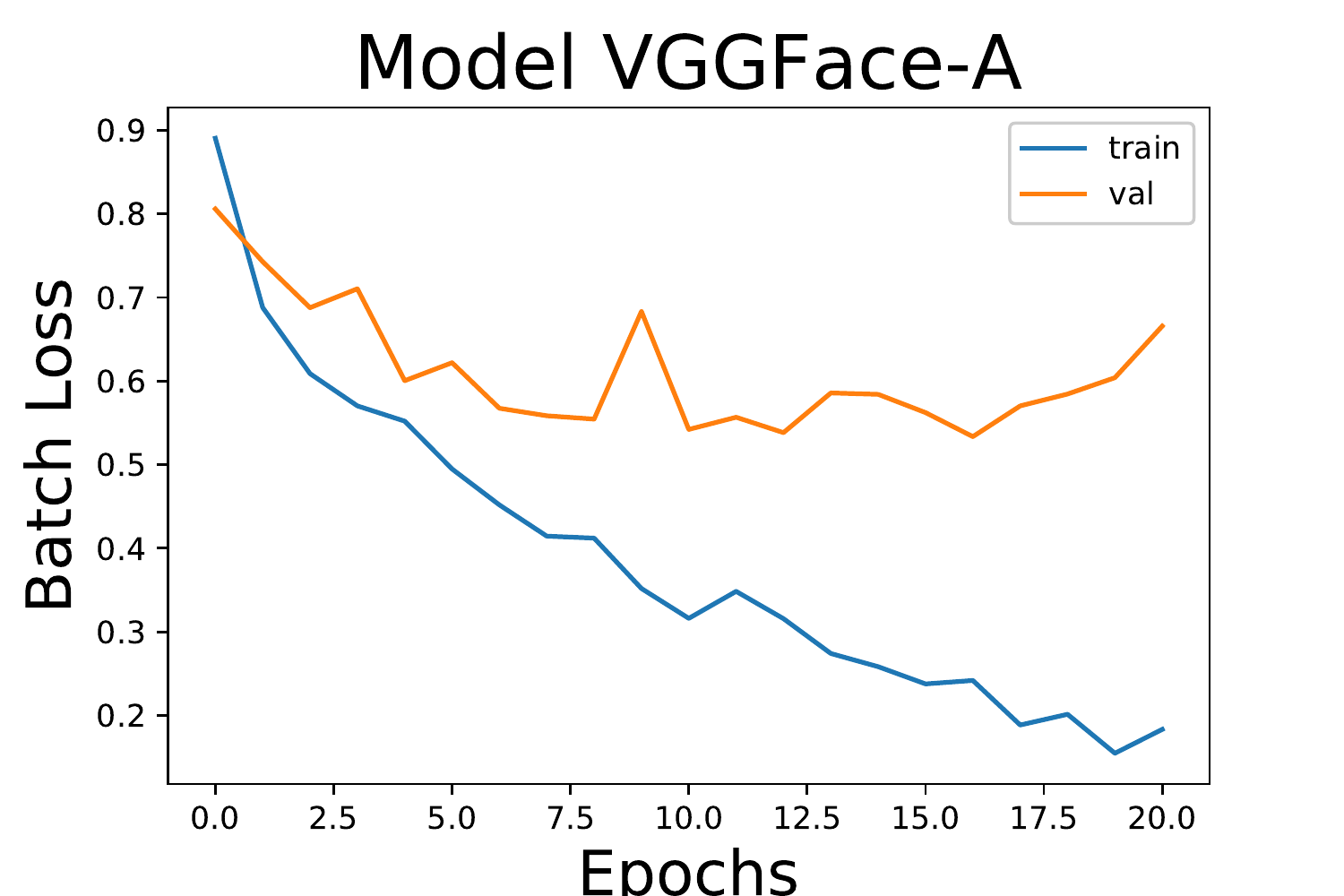}
		\end{subfigure}%
		\vspace{4pt}
		\begin{subfigure}{0.45\linewidth}
			\centering
			\includegraphics[width=\linewidth]{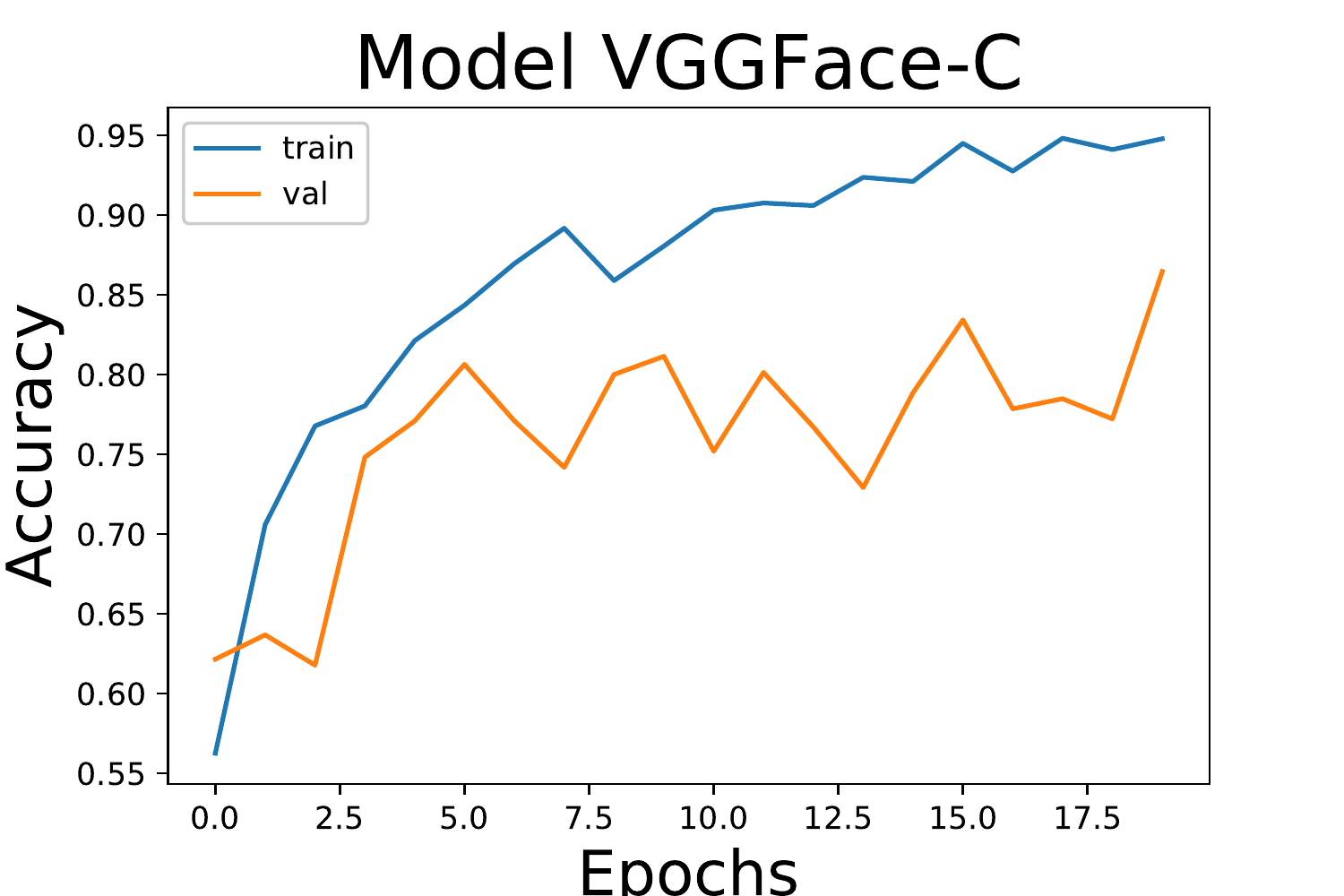}
			\caption{Accuracy Curves}
		\end{subfigure}
		\begin{subfigure}{0.45\linewidth}
			\centering
			\includegraphics[width=\linewidth]{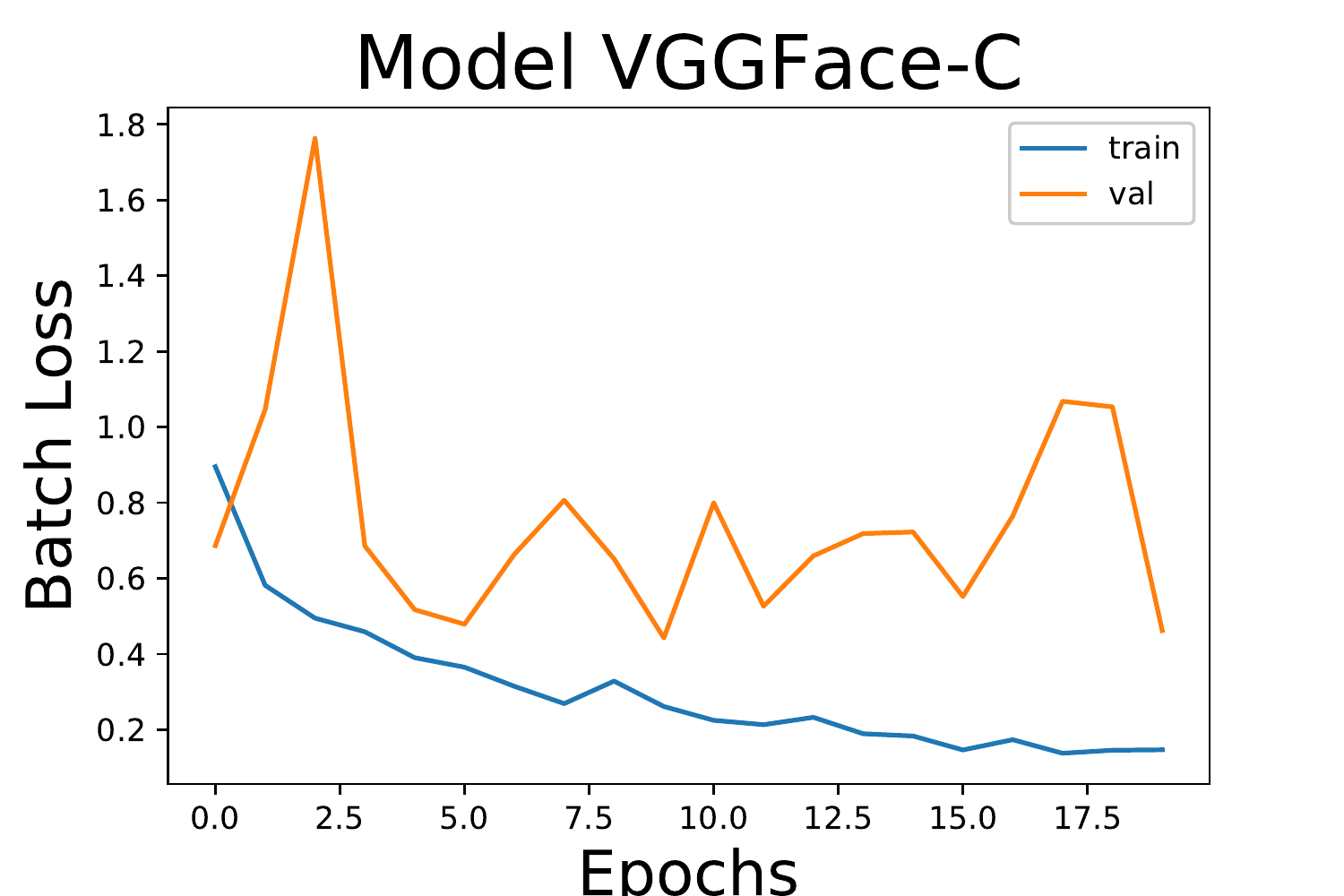}
			\caption{Loss Curves}
		\end{subfigure}	
		\caption{Training and Validation loss and accuracy curves for models VGGFace-A and  VGGFace-C.}
		\label{fig:vggaccloss}
	\end{figure}

\begin{figure}[!t]
		\begin{subfigure}{0.45\linewidth}
			\centering
			\includegraphics[width=\linewidth]{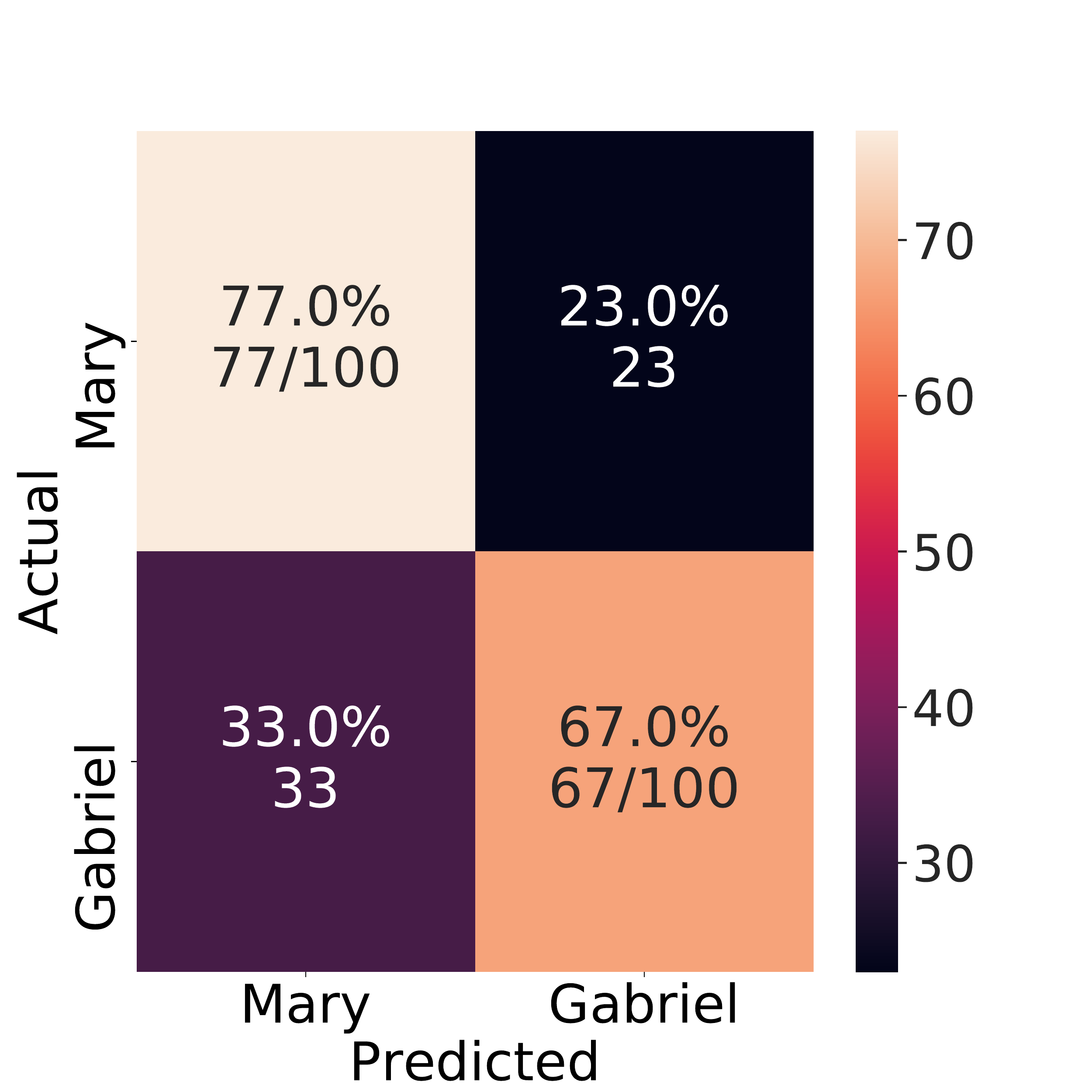}
			\caption{VGGFace-A}
		\end{subfigure}%
		\begin{subfigure}{0.45\linewidth}
			\centering
			\includegraphics[width=\linewidth]{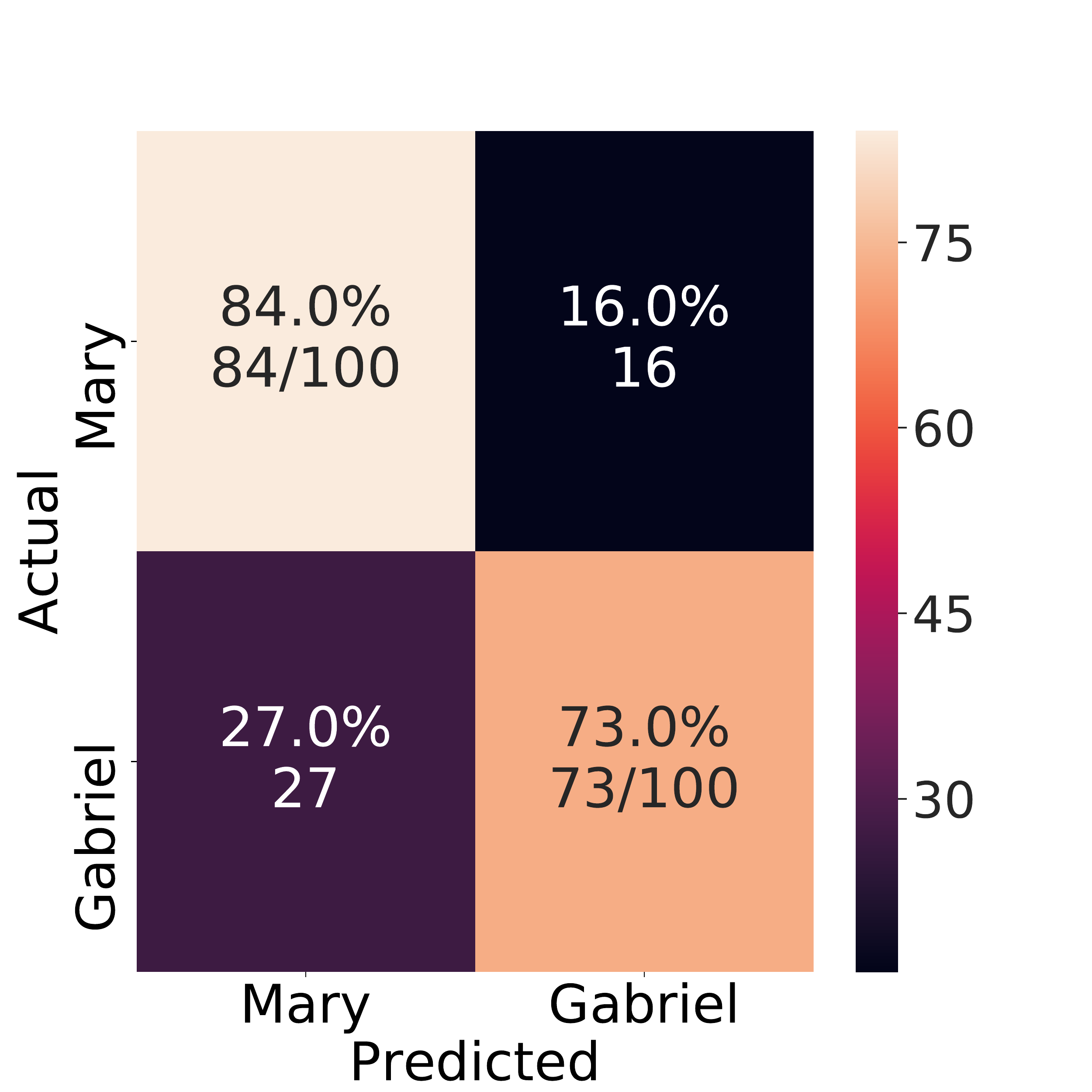}\hfill
			\caption{VGGFace-C}
		\end{subfigure}\par\medskip
		\caption{Confusion Matrices on test set of images of Mary and Gabriel.}
		\label{fig:cms}
	\end{figure}

	\paragraph{\textsc{VGGFace-A:}} Table~\ref{tab:results_all} shows that VGGFace-A outperforms all the traditional ML methods for all the metrics, when trained on body images. It gives comparable performance in comparison to SVM with RBF kernels.  Figure~\ref{fig:vggaccloss} (row~1) shows the accuracy and loss plots during the training of the model. We can see that the training accuracy increases consistently while the validation accuracy stagnates at some point. The model training is stopped when the validation loss stops to improve. We use a tolerance of $0.05$ for validation loss. If the validation loss is lower than the tolerance value for at least $10$ epochs, the model training is halted. We observed empirically that this avoids overfitting the model. The confusion matrix for the model is shown in Figure~\ref{fig:cms}\,(a), where we can see that the model incorrectly predicts $56/200$ test samples for both the classes giving a good quantitative performance. 
	
	\paragraph{\textsc{VGGFace-B:}} This model is a special case, where we fine-tune the VGGFace (ResNet50) directly on the styled dataset. The styled dataset is a combination of images of people (males and females, thereby providing the context for training a two-class classification model) and the styles of the annunciation scenes transferred to those images. Table~\ref{tab:results_all} shows the performance metrics of this model on the test set of images of Mary and Gabriel, which is inferior to that of VGGFace-A. This makes sense since the model has not seen the actual images of Mary and  Gabriel, only their styles. When tested on the test set of styled images, the performance metrics are: \emph{Pr: 0.73, Re: 0.82, F1: 0.77, Acc: 0.76}, which are even better than using VGGFace-A~(Table~\ref{tab:results_all} (row~5)). These metrics suggest that the model has adapted quite well on the styled data due to their larger database size.  
	
	\paragraph{\textsc{VGGFace-C:}} Motivated by the performance of the previous models, we take \emph{VGGFace-B} as our base model and fine-tune it on the Mary and Gabriel dataset. We take the standard parameters and train it until the validation loss saturates. Figure~\ref{fig:vggaccloss} (row~2) shows the accuracy and loss progression over the training epochs for the model. Similar to VGGFace-A, we stop the training when the validation loss does not improve. Here as well we consider the tolerance value for validation loss to be $0.05$, so if the validation loss is within this tolerance value for at least $10$ epochs, then the training is stopped to avoid overfitting. Table~\ref{tab:results_all} shows that this model outperforms all other models for all the metrics. Figure~\ref{fig:cms}\,(b), shows the confusion matrix for this model, we see that the model mis-classified $43/200$ samples in contrast to $56/200$ samples in case of VGGFace-A. We can conclude that the network is able to make use of the domain knowledge acquired through the training on the styled images.
	
	\subsection{CAM Analysis on CNN models}
	\begin{figure}[!t]
		\centering
		%% Gabriel
		\begin{subfigure}[t]{0.24\linewidth}
			\centering
			\includegraphics[width=4cm, height=3cm, keepaspectratio]{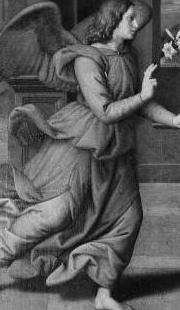}
		\end{subfigure}
		\begin{subfigure}[t]{0.24\linewidth}
			\centering
			\includegraphics[width=4cm, height=3cm, keepaspectratio]{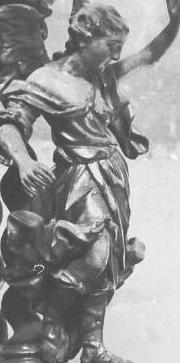}
		\end{subfigure}
		\begin{subfigure}[t]{0.24\linewidth}
			\centering
			\includegraphics[width=4cm, height=3cm, keepaspectratio]{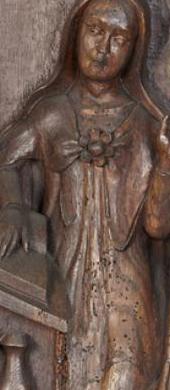}
		\end{subfigure}	
		\begin{subfigure}[t]{0.24\linewidth}
			\centering
			\includegraphics[width=4cm, height=3cm, keepaspectratio]{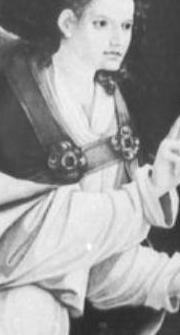}
		\end{subfigure}\vspace{4pt}%
		
		\begin{subfigure}[t]{0.24\linewidth}
			\centering
			\includegraphics[width=4cm, height=3cm, keepaspectratio]{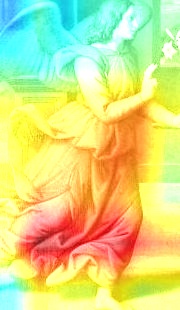}
		\end{subfigure}
		\begin{subfigure}[t]{0.24\linewidth}
			\centering
			\includegraphics[width=4cm, height=3cm, keepaspectratio]{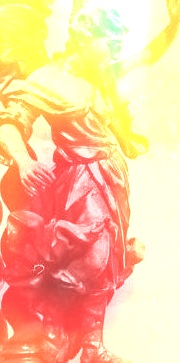}
		\end{subfigure}	
		\begin{subfigure}[t]{0.24\linewidth}
			\centering
			\includegraphics[width=4cm, height=3cm, keepaspectratio]{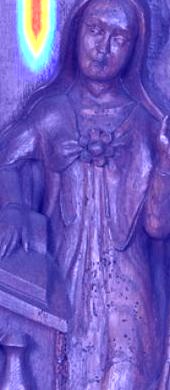}
		\end{subfigure}
		\begin{subfigure}[t]{0.24\linewidth}
			\centering
			\includegraphics[width=4cm, height=3cm, keepaspectratio]{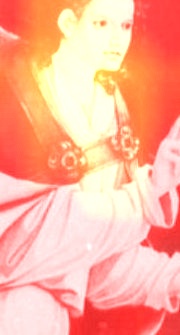}
		\end{subfigure}\vspace{4pt}%

		\begin{subfigure}[t]{0.24\linewidth}
			\centering
			\includegraphics[width=4cm, height=3cm, keepaspectratio]{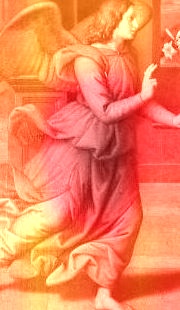}
			\caption{}
		\end{subfigure}
		\begin{subfigure}[t]{0.24\linewidth}
			\centering
			\includegraphics[width=4cm, height=3cm, keepaspectratio]{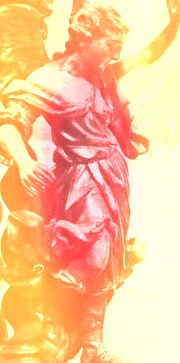}
			\caption{}
		\end{subfigure}
		\begin{subfigure}[t]{0.24\linewidth}
			\centering
			\includegraphics[width=4cm, height=3cm, keepaspectratio]{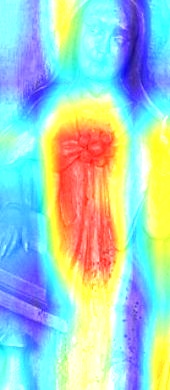}
			\caption{}
		\end{subfigure}
		\begin{subfigure}[t]{0.24\linewidth}
			\centering
			\includegraphics[width=4cm, height=3cm, keepaspectratio]{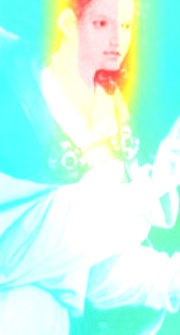}
			\caption{}
		\end{subfigure}
		\caption{\emph{Columns (a, b)} are positively predicted samples of Gabriel from the test set. \emph{Columns (c, d)} are positively predicted samples of Mary from the test set. \emph{Row~1} shows the original images, \emph{Row~2} shows the CAMs generated for model VGGFace-A, whereas \emph{Row~3} shows the CAMs generated for model VGGFace-C.}
		\label{fig:cams}
	\end{figure}
	Visualization of the deep networks has become one of the most important and interesting aspects of deep learning, mainly to understand the internal working of the networks~\cite{zeiler2014visualizing}. Deep CNN networks are powerful enough to learn the localization of objects in higher layers~\cite{zhou2016learning}. \emph{Class Activation Maps} (CAMs)~\cite{zhou2016learning} are an interesting way to visualize the image regions where the deep network is focusing while making the predictions about it's class.
	
	Figure~\ref{fig:cams} shows the CAMs for positively predicted test samples of Mary and Gabriel for VGGFace-A (row~2) and VGGFace-C (row~3) models. We can see how the network tries to localize at the body (including the dress), wings and the neighboring information while predicting for Gabriel and Mary. The results from the CAMs strengthen our argument as to why VGGFace-C is a better model as compared to VGGFace-A. For Figure~\ref{fig:cams}(a, b), VGGFace-A uses features located at the lower part of the clothing, whereas VGGFace-C perceives the whole body. This also shows that the use of the whole body for recognizing character gives more context for the model to learn about the character. In Figure~\ref{fig:cams}(c) VGGFace-C again looks at the body part of Mary for making the prediction, whereas for Figure~\ref{fig:cams}(d) it focuses on the facial area, as opposed to VGGFace-A which looks at other regions of the images except the face. Interesting to note here that both the models are making the correct predictions by looking at different regions. 
	
	\begin{figure}[!t]
		\centering
		%% Gabriel
		\begin{subfigure}[t]{0.24\linewidth}
			\centering
			\includegraphics[width=4cm, height=3cm, keepaspectratio]{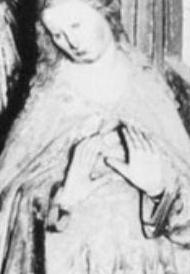}
			\caption{}
		\end{subfigure}
		\begin{subfigure}[t]{0.24\linewidth}
			\centering
			\includegraphics[width=4cm, height=3cm, keepaspectratio]{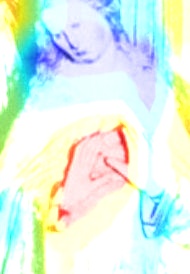}
			\caption{}
		\end{subfigure}
		\begin{subfigure}[t]{0.24\linewidth}
			\centering
			\includegraphics[width=4cm, height=3cm, keepaspectratio]{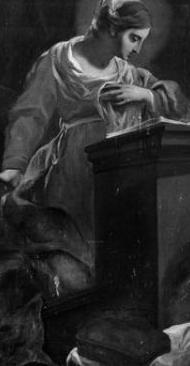}
			\caption{}
		\end{subfigure}
		\begin{subfigure}[t]{0.24\linewidth}
			\centering
			\includegraphics[width=4cm, height=3cm, keepaspectratio]{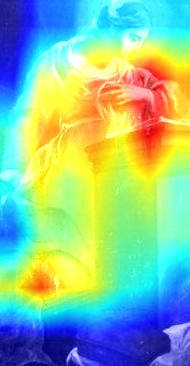}
			\caption{}
		\end{subfigure}
		\caption{(a) and (c) are the original images of Mary from the test set and \emph{(b), (d)} are their corresponding CAMs.}
		\label{fig:semantic_cams}
	\end{figure}
	
	A very peculiar information that the network tries to capture is observed in Figure~\ref{fig:semantic_cams}, where it has correctly classified the class as Mary, but in this case, the focus of the network is on the hand positions of Mary. This is a very useful semantic information that the network is trying to capture when it is provided domain related knowledge.

	\section{Discussions and Conclusion}\label{sec:conclude}
	We demonstrated that the traditional ML techniques are insufficient to learn complex and more contextual features present in the bodies of the characters. Specifically for images from artworks, it is important to visualize more context (body as opposed to only face) to allow for a better analysis of the characters. We showed that styles from one domain can be transferred to another, and this information from styled images is beneficial for improving the performance of the deep CNN models. By looking at the CAMs, we were able to see how these networks capture semantic information present in the annunciation scenes to make the predictions. 
	
	Recognizing characters in art history is a complex problem given the diversity of ways in which the protagonists can be described. We demonstrated that deep CNN models are able to learn the required domain knowledge for recognizing character using style-transferred images. 
	%This technique gives good hopes for transfer of domain knowledge. VC: ?
	This technique allows that available datasets can be style-transferred and then used to fine-tune models before training on the actual datasets, thereby reducing excessive reliance on larger datasets in art history. In the end, it is important to note that using the whole body annotations of Mary and Gabriel, the models are able to perform better since they are able to capture more contextual and semantic information. 
	
	%%%%%%%%%%%%%%%%%%%%%%%%%%%%%%%%%%%%%%%%%%%%%%%%%%%%%%%%%%%%%%%%%%
	\begin{acks}
		This work was funded by Emerging Fields Project~\textit{ICONOGRAPHICS} within the Framework of the FAU\footnote{Friedrich-Alexander-Universität Erlangen-Nürnberg, Germany} Emerging Fields Initiative. This work has also been partly supported by the Cross-border Cooperation Program Czech Republic -- Free State of Bavaria ETS Objective 2014-2020
		(project no. 211). We would also like to thank NVIDIA for their generous hardware donations. 
	\end{acks}

	\bibliographystyle{ACM-Reference-Format}
	\bibliography{main}
\end{document}